\pdfoutput=1 
\documentclass[conference]{IEEEtran}

\usepackage{times}

\usepackage{amsmath,amsfonts,bm}

\usepackage{xcolor}
\usepackage{colortbl}

\def\eqref#1{equation~\ref{#1}}

\def\1{\bm{1}}

\DeclareMathAlphabet{\mathsfit}{\encodingdefault}{\sfdefault}{m}{sl}
\SetMathAlphabet{\mathsfit}{bold}{\encodingdefault}{\sfdefault}{bx}{n}

\def\gJ{{\mathcal{J}}}

\def\gL{{\mathcal{L}}}

\def\gR{{\mathcal{R}}}

\def\sl{{\mathcal{l}}}

\newcommand\bigO{\mathcal{O}}

\newcommand{\dt}{\Delta t}

\usepackage[colorlinks=true,
citecolor=blue,
filecolor=black,
linkcolor=blue,
urlcolor=blue]{hyperref}
\usepackage{url}
\usepackage{graphicx}

\usepackage{multirow}
\usepackage{booktabs}
\usepackage{paralist}

\usepackage{booktabs} 
\usepackage{multicol}

\usepackage[ruled,noend]{algorithm2e}

\SetCommentSty{mycommfont}

\usepackage{here}
\usepackage{graphicx}
\usepackage{caption}
\usepackage{amsmath,amssymb,amsfonts,amsbsy,amsfonts,latexsym}
\usepackage{multirow}
\usepackage{makecell}
\usepackage[labelfont=bf,textfont=it,belowskip=0pt,aboveskip=5pt,tableposition=top]{caption}
\usepackage{xcolor}
\usepackage{colortbl}

\definecolor{colorA}{RGB}{189,201,225}
\definecolor{colorB}{RGB}{103,169,207}
\definecolor{colorC}{RGB}{ 28,144,153}
\definecolor{colorD}{RGB}{  1,108, 89}

\newcolumntype{R}{>{\columncolor{gray!40}}r}
\newcolumntype{L}{>{\columncolor{gray!40}}l}
\newcolumntype{C}{>{\columncolor{gray!40}}c}

\usepackage{tabularx,colortbl,xcolor}
\usepackage{multirow}
\usepackage[normalem]{ulem}
\useunder{\uline}{\ul}{}

\usepackage{enumitem}

\usepackage{xparse}

\graphicspath{{fig}}
\DeclareGraphicsExtensions{.pdf,.png}
\usepackage{pgfplots, pgfplotstable}

\SetKwInput{KwInput}{Input}

\NewDocumentCommand{\var}{O{s} m O{}}{%
  \ensuremath{#1_{#2}^{#3}}
}
\usepackage{siunitx}

\definecolor{light-gray}{gray}{0.80}

\renewcommand\paragraph{\subsubsection*}

\newcommand\eref{Eq.~\ref}
\newcommand\fref{Fig. ~\ref}

\def\0{{\bf 0}}

\usepackage{amsthm}

\newtheorem{mytheorem}{Theorem}
\newtheorem{assumption}[mytheorem]{Assumption}
\newtheorem{lemma}[mytheorem]{Lemma}
\newtheorem{remk}[mytheorem]{Remark}

\graphicspath{{figures/}}

\newcommand{\OURS}[0]{\textsc{ANODE}\xspace}

\usepackage[maxbibnames=99,style=numeric,natbib=true,backend=bibtex, sortcites=true]{biblatex}
\bibliography{ref}

\title{\OURS: Unconditionally Accurate Memory-Efficient Gradients for Neural ODEs}

\author{
Amir Gholami$^1$, Kurt Keutzer$^1$, George Biros$^2$\\
$^1$ Berkeley Artificial Intelligence Research Lab, EECS, UC Berkeley\\
$^2$ Oden Institute, UT Austin
}

\begin{document}
\maketitle
\begin{abstract}
Residual neural networks can be viewed as the forward Euler discretization of an Ordinary
  Differential Equation (ODE) with a unit time step.  This has recently motivated researchers to explore other discretization approaches and train ODE based networks.
However, an important challenge of neural ODEs 
is their prohibitive memory cost during gradient backpropogation.
Recently a method proposed in~\cite{chen2018neural},
claimed that this memory overhead can be reduced
from $\bigO(LN_t)$, where $N_t$ is the number of time steps, down to
$\bigO(L)$ by solving forward ODE backwards in time, where $L$ is the
depth of the network. However, we will show that this approach may
lead to several problems: (i) it may be numerically
unstable for ReLU/non-ReLU activations and general convolution operators,
and (ii) the proposed optimize-then-discretize approach may lead to
divergent training due to inconsistent gradients for small time step sizes. We
discuss the underlying problems, and to address them we propose \OURS, an Adjoint based Neural ODE framework which avoids
the numerical instability related problems noted above, and provides unconditionally
accurate gradients. \OURS has a
memory footprint of $\bigO(L) + \bigO(N_t)$, with the same
computational cost as reversing ODE solve.
We furthermore, discuss a memory efficient
algorithm which can further reduce this footprint with a trade-off of
additional computational cost. We show results on Cifar-10/100
datasets using ResNet and SqueezeNext neural networks.
\end{abstract}

\section{Introduction}
\label{sec:intro}

\begin{figure}[!htbp]
\begin{center}
  \includegraphics[width=.485\textwidth]{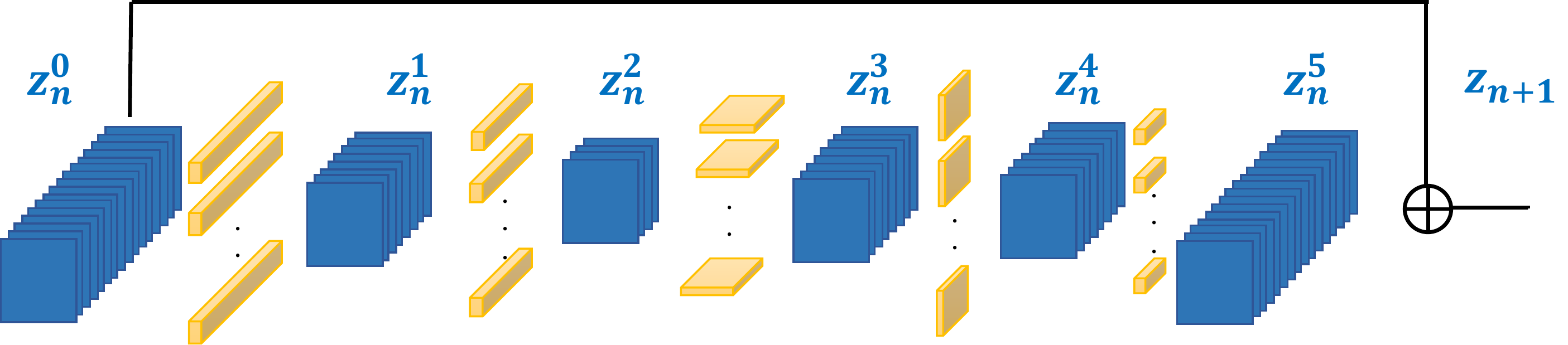}\\\vspace{0.7cm}
  \includegraphics[width=.485\textwidth]{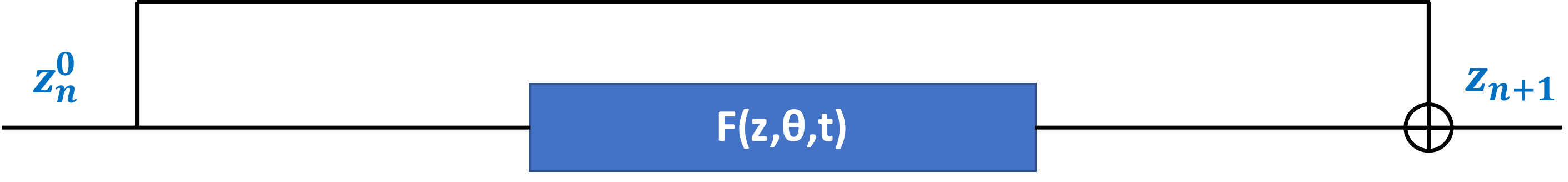}
\end{center}
\caption{A residual block of SqueezeNext~\cite{gholami2018squeezenext} is shown. The input
activation is denoted by $z_n$, along with the intermediate values of $z_n^1,\cdots,z_n^5$.
The output activation is denoted by $z_{n+1}=z_n^5+z_n$. In the second row, we show a compact
representation, by denoting the convolutional blocks as $f(z)$. This residual
block could be thought of solving an ODE~\eref{e:ode} with an Euler scheme.
}
\label{fig:resnet_ode}
\end{figure}

The connection between residual networks and ODEs has been discussed  in~\cite{weinan2017proposal,haber2017stable,ruthotto2018deep,lu2017beyond,ciccone2018nais}.
Following~\fref{fig:resnet_ode}, we define $z_0$  to be the input to a residual net block and $z_1$
its output activation.  Let $\theta$ be the weights and  $f(z,\theta)$ be the nonlinear operator defined by this neural
network block. In general, $f$ comprises a combination of convolutions, activation functions,  and batch normalization operators. For
this residual block we have $z_1 = z_0 + f(z_0,\theta)$. Equivalently we can define
an autonomous ODE $\frac{dz}{dt} = f(z(t),\theta)$ with $z(t=0) = z_0$ and define its output $z_1 = z(t=1)$, where $t$ denotes time. 
The relation to a residual network is immediate if we consider a forward Euler discretization using one step, $z_1 = z_0 + f(z(0),\theta)$.
In summary we have:
\begin{subequations}
\begin{align}
        z_1  &= z_0 + f(z_0,\theta) &&\quad \mbox{\small ResNet} \\
        z_1 &= z_0 + \int_0^1 f(z(t),\theta) dt && \quad \mbox{\small ODE}\label{e:ode} \\
        z_1 &= z_0 + f(z_0,\theta) && \quad \mbox{\small ODE forward Euler}
\end{align}
\end{subequations}

\begin{figure}[t]
\begin{center}
  \includegraphics[width=0.4\textwidth]{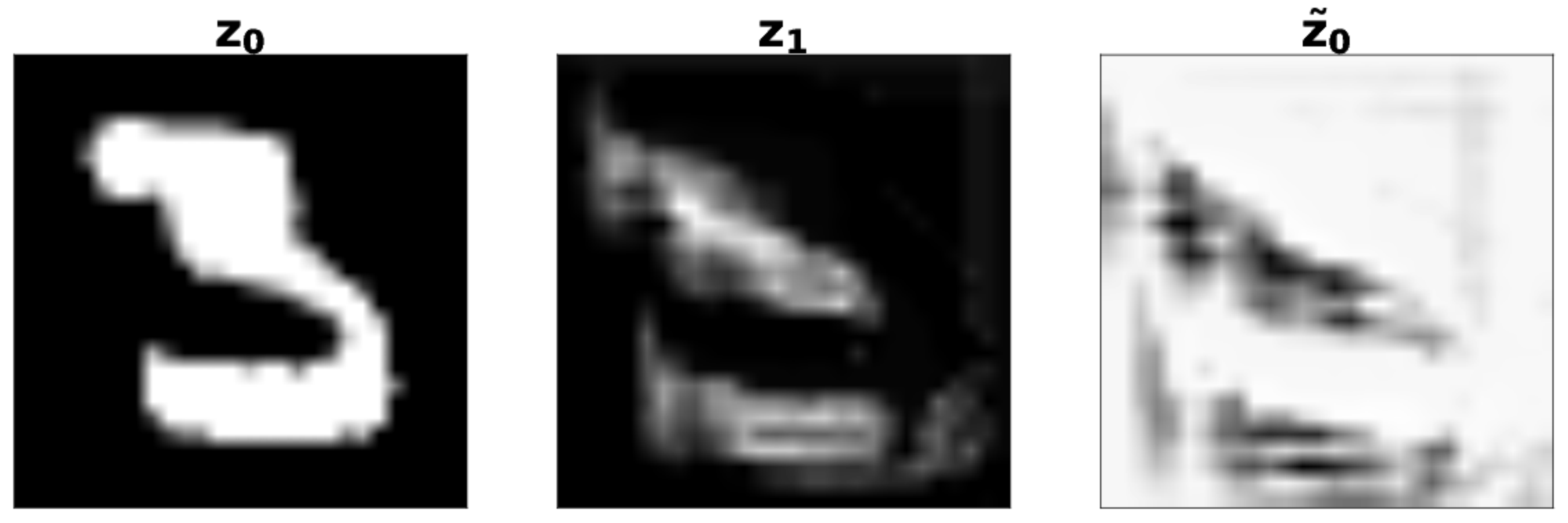}\\
  \includegraphics[width=0.4\textwidth]{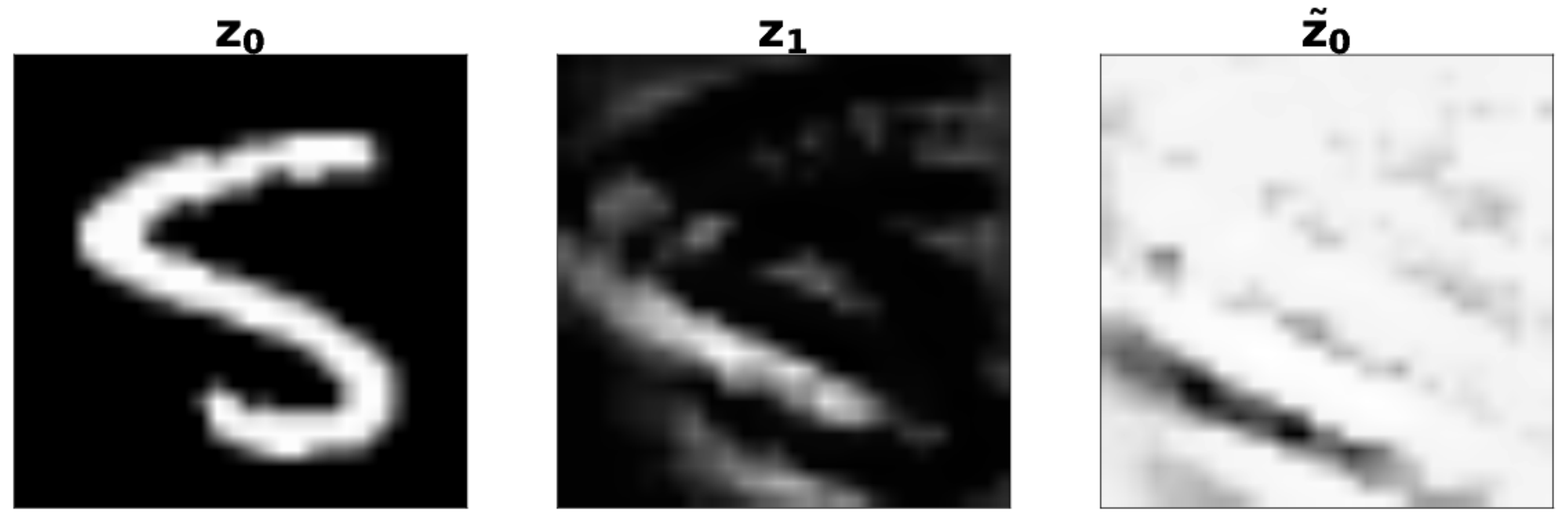}
 \end{center}
\caption{Demonstration of numerical instability associated with reversing
NNs. We consider a single residual block consisting of one convolution layer, with random Gaussian initialization, followed by a ReLU.
The first column shows input image that is fed to the residual block,
results of which is shown in the second column. The last column shows the result when solving the forward problem backwards
as proposed by~\cite{chen2018neural}.
One can clearly see that the third column is completely
different than the original image shown in the first column.
The two rows correspond to ReLU/Leaky ReLU activations, respectively.
Please see~\fref{fig:extra_results_adaptive} for more results with adaptive solvers and with different activation functions. All of the results point
to numerical instability associated with backwards ODE solve.
}
\label{fig:relu_mnist_example}
\end{figure}

This new ODE-related perspective can lead to new insights regarding stability and trainability of networks, as well as new network architectures inspired by ODE discretization schemes. However, there is a major challenge in implementation of neural ODEs. Computing the gradients of an ODE layer through backpropagation requires
storing all intermediate ODE solutions in time (either through auto differentiation or an adjoint method~\S\ref{subsec:adjoint}), which can easily lead to  prohibitive memory costs. For example, if we were to use two forward Euler time steps for~\eref{e:ode}, we would have $z_{1/2}=z_0 + 1/2 f(z_0,\theta)$ and $z_1 = z_{1/2} + 1/2 f(z_{1/2},\theta)$. We need to store all $z_0$, $z_{1/2}$ and $z_1$ in order to compute the gradient with respect the model weights $\theta$. Therefore, if we take $N_t$ time steps then we require $\bigO(N_t)$ storage per ODE block. If we have $L$ residual blocks, each being a separate ODE, the storage requirements can increase from $\bigO(L)$ to $\bigO(L N_t)$.

In the recent work of~\cite{chen2018neural}, an adjoint based backpropagation 
method was proposed to address this challenge; it only requires $\bigO(L)$ memory to compute the gradient with respect $\theta$, thus significantly outperforming existing backpropagation implementations.  The basic idea is that instead of  storing $z(t)$, we store only $z(t=1)$
and then we \emph{reverse the ODE solver in time} by  solving $\frac{dz}{dt} = -f(z(t),\theta)$, in order to reconstruct $z(t=0)$ and evaluate the gradient.
However, we will show that using this approach does not work for any value $\theta$ for a general NN model. It may lead to significant ($\bigO(1)$) errors in the gradient
because reversing the ODE may not be possible and, even if it is, numerical discretization errors may cause the gradient to diverge.

Here we discuss the underlying issues and then present \OURS, an Adjoint based Neural ODE framework that employs a classic \emph{``checkpointing''} scheme that addresses the memory problem and results in correct gradient calculation no matter the structure of the underlying network.  In  particular, we make the following contributions:
\begin{itemize}
    \item We show that neural ODEs with ReLU activations may not be reversible (see~\S\ref{sec:flows}). Attempting to solve such ODEs backwards in time, so as to recover activations $z(t)$ in earlier layers, can lead to mathematically incorrect gradient information. This in turn may lead to divergence of the training.  This is illustrated in Figs.~\ref{fig:relu_mnist_example},~\ref{fig:dto_vs_otd},~\ref{fig:resnet_dto_vs_otd}, and~\ref{fig:extra_results_adaptive}.
    
    \item We show that even for general convolution/activation operators, the reversibility of neural ODE may be numerically unstable. This is due to possible ill-conditioning of the reverse ODE solver, which can amplify numerical noise over large time horizons (see~\S\ref{sec:flows}). An example of instability is shown in~\fref{fig:relu_mnist_example}.
    Note that this instability cannot be resolved by using adaptive time stepping as shown in~\fref{fig:extra_results_adaptive}.
    
    \item We discuss an important consistency problem between discrete gradient and continuous gradient,
    and show that ignoring this inconsistency can lead to divergent training. We illustrate that this problem stems
    from well established issue related to the difference between \emph{``Optimize-Then-Discretize''} differentiation methods versus \emph{``Discretize-Then-Optimize''} differentiation methods (see~\S\ref{sec:otd_vs_dto}).
    
    \item We present \OURS, which is a neural ODE framework with checkpointing that uses
    \emph{``Discretize-Then-Optimize''} differentiation method (see~\S\ref{sec:sqode}). \OURS avoids
    the above problem, and squeezes the memory footprint to 
    $\bigO(L) + \bigO(N_t)$ from $\bigO(LN_t)$, without suffering from the above problems. 
    This footprint can be further reduced with additional computational cost
    through logarithmic checkpointing methods.
    Preliminary tests show efficacy of our approach. A comparison between \OURS and neural ODE~\cite{chen2018neural} is shown in~\fref{fig:dto_vs_otd},~\ref{fig:resnet_dto_vs_otd}.
\end{itemize}

\subsection{Related Work}
\textbf{ODEs:} In~\cite{weinan2017proposal}, the authors observed that residual networks can be viewed as a forward Euler discretization of an ODE. The stability of the forward ODE problem was discussed in~\cite{haber2017stable,chang2018reversible}, where the authors proposed architectures that are discrete Hamiltonian systems and thus are both stable and reversible in both directions. In an earlier work, a similar reversible 
architecture was proposed for unsupervised generative models~\cite{dinh2014nice,dinh2016density}.  In~\cite{gomez2017reversible}, a reversible architecture was used to design a residual network to avoid storing intermediate activations.

\noindent
\textbf{Adjoint Methods:} Adjoint based methods
have been widely used to solve problems ranging from classical optimal control problems~\cite{lions1971optimal} and
large scale inverse problems~\cite{biros2005parallel,biros2005parallel2}, to
climate simulation~\cite{charpentier2000efficient,lafore1997meso}.
Despite their prohibitive memory footprint, adjoint methods
were very popular in the nineties, mainly due to the computational efficiency for 
computing gradients as compared to other methods.
However, the limited available memory at the time, prompted researchers to explore
alternative approaches to use adjoint without having to store all of the forward
problem's trajectory in time. For certain problems such as climate simulations, this
is still a major consideration, even with the larger memory available today.
A seminal work that discussed an efficient solution to overcome this memory overhead
was first proposed in~\cite{griewank1992achieving}, with a comprehensive analysis in~\cite{griewank2000algorithm}. It was proved that for given a fixed memory budget, one
can obtain optimal computational complexity using a binomial checkpointing method.

The checkpointing method was recently \emph{re-discovered} in machine learning for backpropagation
through long recurrent neural networks (NNs) in~\cite{martens2012training}, by checkpointing
a square root of the time steps. During backpropagation, the intermediate activations
were then computed by an additional forward pass on these checkpoints.
Similar approaches to the optimal checkpointing, originally proposed in~\cite{griewank1992achieving,griewank2000algorithm}, were recently explored
for memory efficient NNs in~\cite{chen2016training,gruslys2016memory}.

\section{Problem Description}
\label{sec:analysis}
We first start by introducing some notation. Let us denote the input/output training data as $x\in\gR^d_x$ and $y\in\gR^d_y$, drawn from some unknown probability distribution
$P(x,y): \gR^d_x\times\gR^d_y \rightarrow [0,1]$. The goal is to learn the mapping between
$y$ and $x$ via a model $F(\theta)$, with $\theta$ being the unknown parameters.
In practice we do not have access to the joint probability of $P(x,y)$, and thus
we typically
use the empirical loss over a set of $m$ training examples:

\begin{equation}
    \min_\theta \gJ(\theta, x, y) = \frac{1}{m}\sum_{i=1}^m \ell(F(\theta, x_i), y_i) + \gR(\theta),
\end{equation}
where the last term, $\gR(\theta)$, is some regularization operator such as weight decay.
This optimization problem is solved iteratively, starting with some (random) initialization for $\theta$. Using this initialization for the network, we then 
compute the activation of each layer by solving~\eref{e:ode} forward in time
(along with possibly other transition, pooling, or fully connected layers).
After the forward solve, we obtain $F(z)$, which is the prediction of the NN (e.g. a vector of probabilities for each class). 
We can then compute a pre-specified loss (such as cross entropy) between
this prediction and the ground truth training data, $y$. 
To train the network, we need to compute the gradient of the loss with respect to model parameters, $\theta$. This would require backpropogating the gradient through ODE layers.

A popular iterative scheme for finding optimal value for $\theta$ is Stochastic Gradient Descent (SGD), where
at each iteration a mini-batch of size $B$ training examples are drawn, and the loss is evaluated over this mini-batch, using a variant of the following update equation:

\begin{equation}
    \theta^{\text{new}} = \theta^{\text{old}} - \eta \frac{1}{B}\sum^B_{i=1}\nabla_\theta \ell(F(\theta, x_i), y_i).
    \label{e:sgd}
\end{equation}

Typically the gradient is computed using chain rule by constructing a graph of computations during the forward pass. For ODE based models, the forward operator would involve integration of~\eref{e:ode}. To backpropagate through this integration, we need  to solve a so called \textit{adjoint} equation.
However, as we will demonstrate, the memory requirement can easily become prohibitive even for shallow NNs.

\subsection{Adjoint Based Backpropogation}
\label{subsec:adjoint}

We first demonstrate how backpropogation can be performed for an ODE layer.
For clarity let us consider a single ODE in isolation and denote 
its output activation with $z_1$ which is computed by solving~\eref{e:ode} forward in time.
During backpropogation phase, we are given the gradient of the loss with respect to output, $\frac{\partial \gJ}{\partial z_1}$, and
need to compute two gradients: (i) gradient w.r.t. model parameters $\frac{\partial \gJ}{\partial \theta}$ which will be
used in~\eref{e:sgd} to update $\theta$, and (ii) backpropogate the gradient through the ODE layer and  compute $\frac{\partial \gJ}{\partial z_0}$.
To compute these gradients, we first form the Lagrangian for this problem defined as:

\begin{align}
    \gL = \gJ(z_1, \theta) + \int_0^1\alpha(t) \cdot \left(\frac{dz}{dt} - f(z,\theta)\right)dt,
\end{align}
where $\alpha$ is the so called ``adjoint'' variable. Using the Lagrangian removes the ODE constraints and the corresponding first
order optimality conditions could be found by taking variations w.r.t. state, adjoint, and NN parameters (the so called Karush-Kuhn-Tucker (KKT) conditions):

\[\arraycolsep=1.4pt\def\arraystretch{2.2}
\left\{
  \begin{array}{@{}lll@{}}
\displaystyle{\frac{\partial \gL}{\partial z} }&=0 \quad\Rightarrow\quad \ \textrm{adjoint equations \  }\\
\displaystyle{\frac{\partial \gL}{\partial \theta}} &=0 \quad \Rightarrow\quad \ \textrm{inversion equation}\\
\displaystyle{\frac{\partial \gL}{\partial \alpha}} &=0  \quad\Rightarrow\quad \ \textrm{state equations \ \ \ }
  \end{array}\right.
\]

This results in the following system of ODEs:

\begin{subequations}
  \begin{align}
    \frac{\partial z}{\partial t} + f(z, \theta) = 0, \quad t \in (0,1] \label{e:kkt_z}\\ 
    -\frac{\partial \alpha(t)}{\partial t} - \frac{\partial f}{\partial z}^T \alpha = 0, \quad t \in [0,1)\label{e:kkt_alpha}\\ 
    \alpha_1 + \frac{\partial J}{\partial z_1} = 0, \label{e:kkt_alpha_1}\\
    g_\theta = \frac{\partial R}{\partial \theta}  - \int_0^1 \frac{\partial f}{\partial \theta}^T\alpha\label{e:kkt_theta}
  \end{align}
\end{subequations}

For detailed derivation of the above equations please see Appendix~\S\ref{s:OTD}. Computation of the gradient follows these steps.
We first perform the forward pass by solving~\eref{e:kkt_z}. Then during backpropogation we are given the gradient w.r.t.
the output activation, i.e., $\frac{\partial J}{\partial z_1}$, which can be used to compute $\alpha_1$ from~\eref{e:kkt_alpha_1}.
Using this terminal value condition, we then need to solve the adjoint ODE~\eref{e:kkt_alpha} to compute $\alpha_0$ which is 
equivalent to backpropogating gradient w.r.t. input. Finally the gradient w.r.t. model parameters could be computed by plugging
in the values for $\alpha(t)$ and $z(t)$ into~\eref{e:kkt_theta}.

It can be clearly seen that solving either~\eref{e:kkt_alpha}, or~\eref{e:kkt_theta} requires 
knowledge of the activation throughout
time, $z(t)$. Storage of all activations throughout this trajectory in time  leads to
a storage
complexity that scales as $\bigO(LN_t)$,
where $L$ is the depth of the network (number of ODE blocks) and $N_t$ is the number of time steps (or intermediate 
discretization values) which was used
to solve~\eref{e:ode}.
This can quickly become very expensive even for shallow NNs,
which has been the main challenge in deployment of neural ODEs.

In the recent work of~\cite{chen2018neural}, an
idea was presented to
reduce this cost down to $\bigO(L)$ instead of
$\bigO(LN_t)$. The method is based on the assumption that the activation functions $z(t)$ can be computed by 
solving the forward ODE backwards (aka reverse) in time.
That is given $z(t=1)$, we can solve~\eref{e:ode} backwards, to obtain intermediate values of the activation function.
However, as we demonstrate below this method may lead to 
incorrect/noisy gradient information for general NNs.


\section{Can we reverse an ODE?}\label{sec:flows}
Here we summarize some well-known results for discrete and continuous dynamical systems.  Let $\frac{dz(t)}{dt} = f(z(t))$, with $z \in \mathbb{R}^n$, 
be an autonomous ODE where $f(z)$ is locally Lipschitz continuous. The Picard-Linderl\"{o}f theorem states that, locally, 
this ODE has a unique solution which depends continuously on the initial condition~\cite{abraham2007manifolds} (page 214).  The \emph{flow} $\phi(z_0,s)$ of this
ODE is defined as its solution with $z(t=0)=z_0$ with time horizon $s$ and is a mapping from $\mathbb{R}^n$ to $\mathbb{R}^n$ and satisfies $\phi(z_0, s+t)= \phi(\phi(z_0,s),t)$. 

If $f$ is $C^1$ and $z_0$ is the initial condition, then there is a
time horizon $I_0:=[0,\tau]$ ($\tau$, which depends on  $z_0$) for which $\phi$ is a \emph{diffeomorphism}, and thus for any $t \in I_0$ $\phi(\phi(z_0,t),-t) = z_0$ \cite{abraham2007manifolds} (page 218).
In other words, this result asserts that if $f$ is smooth enough, then, up to certain time horizon, \emph{which depends on the initial condition}, the ODE is reversible. For example, consider the following ODE with $f \in C^\infty$ : $\frac{dz(t)}{dt} = z(t)^3$, with $z(0)=z_0$. We can verify that the flow of this ODE is given by $\phi(z_0,t)=\frac{z_0}{\sqrt{1-2z_0^2 t}}$, which is only defined for $t< \frac{1}{2 z_0^2}$. 
This simple example reveals a first possible source of problems since the $\theta$ and time horizon (which defines $f$) will be used for all points $z_0$ in the training  set and reversibility is not guaranteed for all of them.

It turns that this \emph{local} reversibility  (i.e., the fact that the ODE is reversible only if the time horizon is less or equal to $\tau$) can be extended to Lipschitz functions~\cite{flow-box}. (In the Lipschitz case, the flow and its inverse are Lipschitz continuous but not diffeomorphic.) This result is of particular interest since the ReLU activation is Lipschitz continuous. Therefore, an  ODE with $f(z) = \max(0, \lambda z)$ (for $\lambda \in \mathbb{R}$) is reversible.  

The reverse flow can be computed by solving $\frac{dz}{ds} = -f(z(s))$ with initial condition $z(s=0) = z(\tau)$ (i.e., the solution of the forward ODE at $t=\tau$). Notice the negative sign in the right hand side of the reverse ODE. \emph{The reverse ODE is not the same as the forward ODE.}   The negative sign in the right hand side of the reverse ODE, reverses the sign of the eigenvalues of the derivative of $f$. If the Lipschitz constant is too large, then reversing the time may create numerical instability that will lead to unbounded errors for the reverse ODE, even if the forward ODE is stable and easy to resolve numerically. To illustrate this consider a simple example. The solution of linear ODEs of the form $dz/dt = \lambda z$, is $z(t) = z_0 \exp(\lambda t)$. If $\lambda<0$, resolving $z(t)$ with a simple solver is easy. But reversing it is numerically unstable (since small errors get amplified exponentially fast).
Consider $\lambda = -100$, i.e., $dz/dt = -100 z$, with $z(0)=1$ and unit time horizon. A simple way to measure the reversibility of the ODE is the following error metric:
\begin{equation}\label{e:error}
\rho(z(0),t) = \frac{\| \phi(\phi(z(0),t),-t)-z(0) \|_2}{\|z(0)\|_2}.
\end{equation}
For $t=1$, Resolving \emph{both}  forward and backward flows up to 1\%    requires about 200,000 time steps (using either an  explicit or an implicit scheme).  For $\lambda=-1e4$, the flow is impossible to reverse numerically in double precision. Although these example might seem to be contrived instabilities, they are actually quite common in dynamical systems. For example, the linear heat equation can result in much larger $\lambda$ values and is not reversible~\cite{fu2007fourier}.  

These numerical issues from linear ODEs extend to ReLU ODEs. Computing $\phi(\phi(z_0,1),-1)$ for $\frac{dz(t)}{dt} = -\max(0, 10 z(t)),\quad z(0)=1, \quad t \in (0,1],$  with $\mathtt{ode45}$ method leads to 1\% error $|\phi(\phi(z(0),1),-1)-z(0)|$ with 11 time steps; and  0.4\% error with 18 time steps. Single machine precision requires 211 time steps using MATLAB's $\mathtt{ode45}$ solver.%
    \footnote{Switching to an implicit time stepping scheme doesn't help.}
 As a second example, consider
    \begin{equation}\label{e:matrelu}
      \frac{dz(t)}{dt} = \max(0, W z(t)),
    \end{equation}
      where $z(t) \in \mathbb{R}^n$ and $W \in \mathbb{R}^{n\times n}$ is a Gaussian random matrix, which is used sometimes as initialization of weights for both fully connected  and convolutional networks. As shown in~\cite{shen2001singular}, $\| W\|_2$ grows as $\sqrt{n}$ and numerically reversing this ODE is nearly impossible for $n$ as small as 100 (it requires 10,000 time steps to get single precision accuracy in error defined by \eqref{e:error}). Normalizing $W$ so that $\|W\|_2 = \bigO(1)$, makes the reversion numerically possible.

A complementary  way to view these problems is by considering the reversibility of discrete dynamical systems. For a repeated map $F^{(n)}  = F \circ F \circ F \circ \cdots \circ F$ the reverse  map can be defined as $F^{(-n)} = F^{-1} \circ F^{-1} \circ F^{-1} \circ \cdots \circ F^{-1}$, which makes it computationally expensive and possibly unstable if $F$ is nearly singular.
For example, consider the map $F(z) = z + a \max(0,z)$ which resembles a single forward Euler time step of a ReLU residual block.
Notice that simply reversing the sign is not sufficient to recover the original input.
It is easy to verify that if $z_0>0$ then $z_0 \neq z_1 - a \max(0, z_1)$ results in error of magnitude $|a^2 z_0|$.
If we want to reverse the map, we need to solve $y + a \max(0, y) = x_1$ for $y$ given $x_1$. Even then this system in general may not have a unique solution (e.g., $-2=y-3 \max(0,y)$ is satisfied for both $y=-2$ and $y=1$). Even if it has a unique solution (for example if $F(z)=z+f(z)$ and $f(z)$ is a contraction~\cite{abraham2007manifolds} (page 124)) it requires a nonlinear solve, which again may introduce numerical instabilities.
In a practical setting an ODE node may involve several ReLUs, convolutions, and batch normalization, and thus its reversibility is unclear.
For example, in NNs some convolution operators resemble differentiation operators, that can lead to very large $\lambda$ values (similar to the heat equation).
A relevant example is a $3\times3$ convolution that is equivalent to the discretization of a Laplace operator. Another example is a non-ReLU activation function such as Leaky ReLU.
We illustrate the latter case in~\fref{fig:relu_mnist_example} (second row). We consider a residual block with Leaky ReLU using an MNIST image as input.
As we can see, solving the forward ODE backwards in time leads to significant errors.
In particular note that this instability cannot be resolved through adaptive time stepping.
We show this for different activation functions in~\fref{fig:extra_results_adaptive}.

\begin{figure*}[t]
\centering
\includegraphics[width=.495\textwidth]{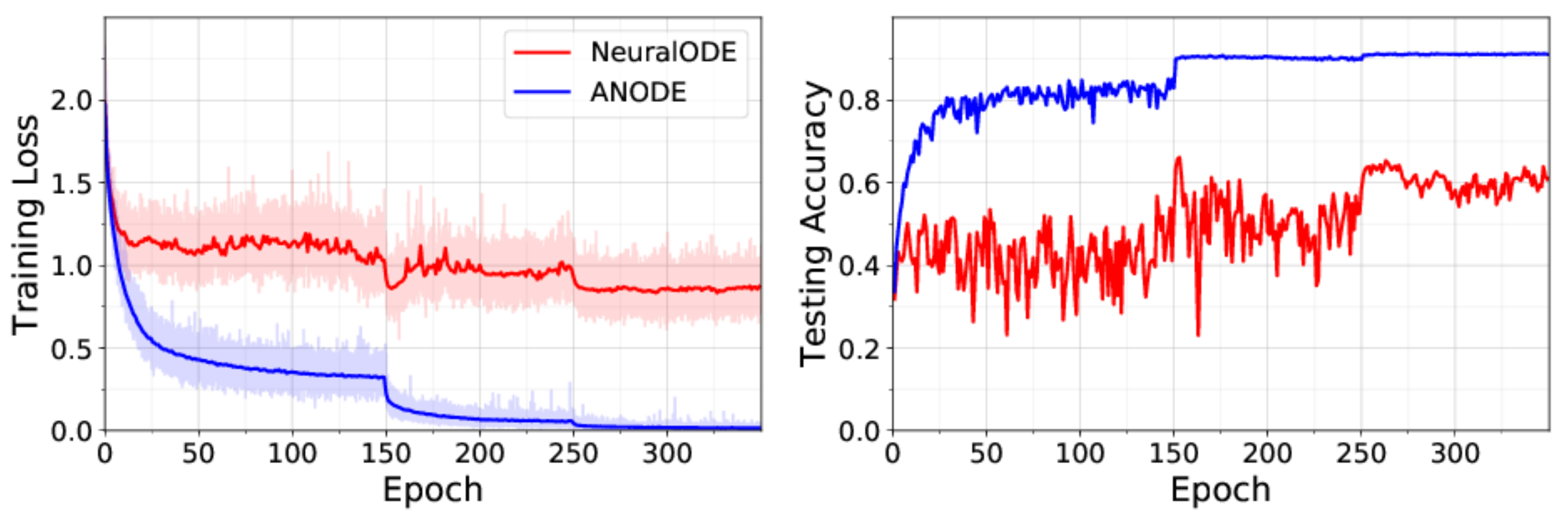}
\includegraphics[width=.495\textwidth]{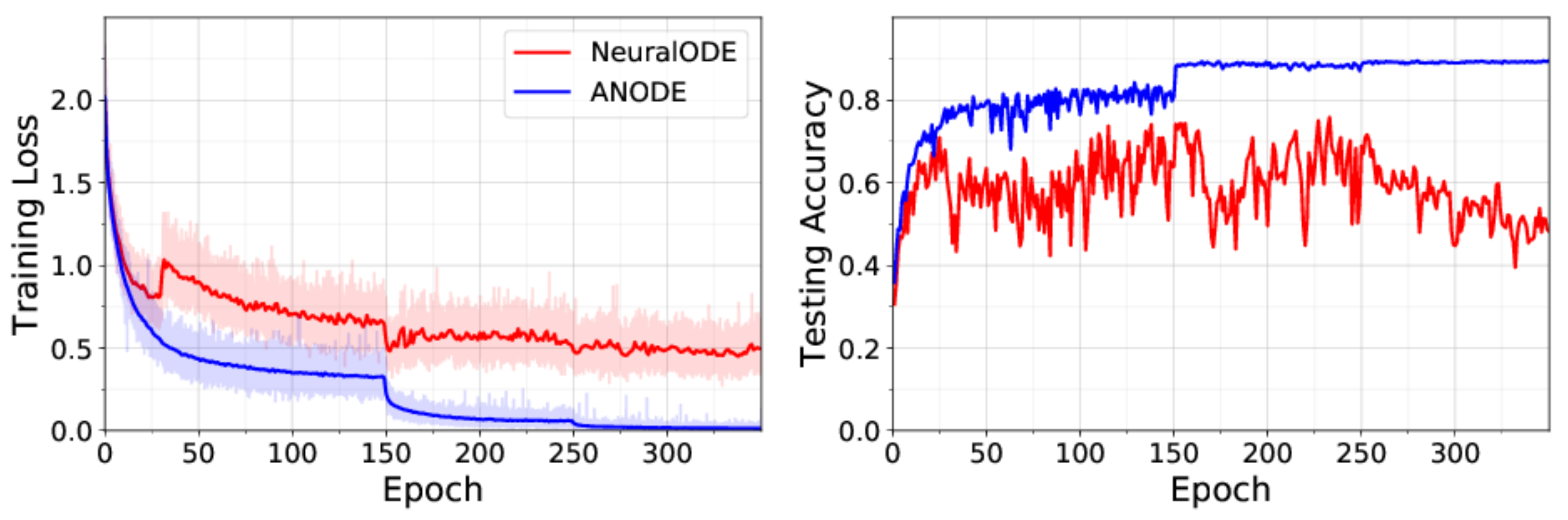}
\caption{Training loss (left) and Testing accuracy (right) for on Cifar-10.
We consider a SqueezeNext network where non-transition blocks are replaced
with an ODE block, solved with Euler method (top) and
RK-2 (Trapezoidal method).
As one can see, the gradient computed using~\cite{chen2018neural} results
in sub-optimal performance, compared to \OURS.
Furthermore, testing~\cite{chen2018neural} with RK45 lead to divergent training in the first epoch.
}
\label{fig:dto_vs_otd}
\end{figure*}

Without special formulations/algorithms, ODEs and discrete dynamical systems can be problematic to reverse. Therefore, computing the gradient (see~\eref{e:kkt_alpha} and~\eref{e:kkt_theta}), using reverse flows may lead to $\bigO(1)$ errors. In~\cite{chen2018neural} it was shown that such approach gives good accuracy on MNIST. However, MNIST is a simple dataset and as we will show, testing a slightly more complex dataset such as Cifar-10 will clearly demonstrates the problem.  It should also be noted that it is possible to avoid this stability problem, but it requires either a large number of time steps or specific NN design~\cite{dinh2014nice,gomez2017reversible,chang2018reversible}.
In particular,  Hamiltonian ODEs and the corresponding discrete systems~\cite{haber2017stable} allow for stable reversibility in \emph{both} continuous and discrete settings. 
Hamiltonian ODEs and discrete dynamical systems are reversible to machine precision as long as appropriate time-stepping is used.
A shortcoming of Hamiltonian ODEs is that so far their performance has not matched the state of the art in standard benchmarks.

In summary, our main conclusion of this section is that activation functions and ResNet blocks that are Lipschitz continuous could be reversible in theory (under certain constraints for magnitude of the Lipschitz constant), but in practice there are several complications due to instabilities and numerical discretization errors. Next, we discuss another issue with computing derivatives for neural ODEs.


\section{Optimize-Then-Discretize versus Discretize-Then-Optimize}
\label{sec:otd_vs_dto}

An important consideration with adjoint methods is the impact of discretization 
scheme for solving the ODEs. For a general function, we can neither solve
the forward ODE of~\eref{e:ode} nor the adjoint ODE of~\eref{e:kkt_alpha} analytically. Therefore, we have to solve these ODEs by approximating
the derivatives using variants of finite difference schemes such as Euler
method.
However, a naive use of such methods can 
lead to subtle inconsistencies which can result in incorrect gradient signal.
The problem arises from the fact that we derive the continuous form
for the adjoint in terms of an integral equation.
This approach is called Optimize-Then-Discretize (OTD).
In OTD method, there is no consideration of the discretization scheme,
and thus there is no guarantee that the finite
difference approximation to the continuous form of the equations would lead to correct gradient information.
We show a simple illustration of the problem by 
considering an explicit Euler scheme with a single time step for solving~\eref{e:ode}:

\begin{equation}
    z_1 = z_0 + f(z_0, \theta).
\end{equation}

During gradient backpropogation we are given $\frac{\partial L}{\partial z_1}$,
and need to compute the gradient w.r.t. input (i.e. $z_0$). 
The \emph{correct} gradient information can be computed through chain rule as follows:
\begin{equation}
    \frac{\partial L}{\partial z_0} = \frac{\partial L}{\partial z_1}(I + \frac{\partial f(z_0, \theta)}{\partial z_0}).
    \label{e:correct_gradient}
\end{equation}

If we instead attempt to compute this gradient with the OTD adjoint method we will have:

\begin{equation}
    \alpha_0 = \alpha_1 (I + \frac{\partial f(z_1, \theta)}{\partial z_1}),
    \label{e:continous_gradient}
\end{equation}
where $a_1=-\frac{\partial L}{\partial z_1}$. It can be clearly seen that the results from DTO approach (\eref{e:correct_gradient}), and OTD (\eref{e:continous_gradient}) can be
completely different. This is because in general $\frac{\partial f(z_1, \theta)}{\partial z_1} \neq \frac{\partial f(z_0, \theta)}{\partial z_0}$.
In fact using OTD's gradient is as if
we backpropogate the gradient by incorrectly replacing input of the neural
network with its output.
Except for rare cases~\cite{gholami_thesis}, the error in OTD and DTO's gradient scales as $\bigO(dt)$. Therefore, this error can become quite large for
small time step sizes.

This inconsistency can be addressed by deriving
discretized optimality conditions, instead of using continuous form.
This approach is commonly referred to as Discretize-Then-Optimize (DTO).
Another solution is to use \textit{self adjoint} discretization
schemes such as RK2 or Implicit schemes. However, the latter methods can be expensive as
they require solving a system of linear
equations for each time
step of the forward operator. For discretization schemes which are
not self adjoint, one needs to solve the adjoint problem derived
from DTO approach, instead of the OTD method.

\begin{figure}[t]
\centering
\includegraphics[width=.495\textwidth]{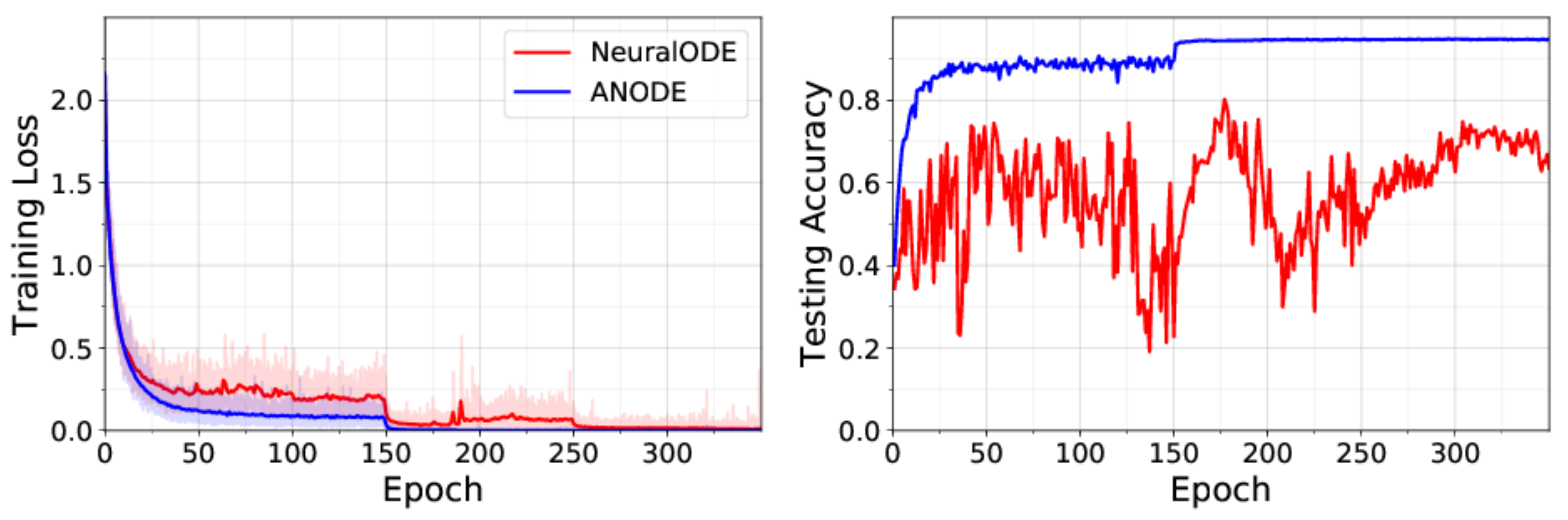}
\caption{Training loss (left) and Testing accuracy (right) for on Cifar-10.
We consider a ResNet-18 network where non-transition blocks are replaced
with an ODE block, solved with Euler method.
As one can see, the gradient computed using~\cite{chen2018neural} results
in sub-optimal performance, compared to \OURS.
Furthermore, testing~\cite{chen2018neural} with RK45 leads to divergent training in the first epoch.
}
\label{fig:resnet_dto_vs_otd}
\end{figure}

\begin{figure}[t]
\centering
\includegraphics[width=.495\textwidth]{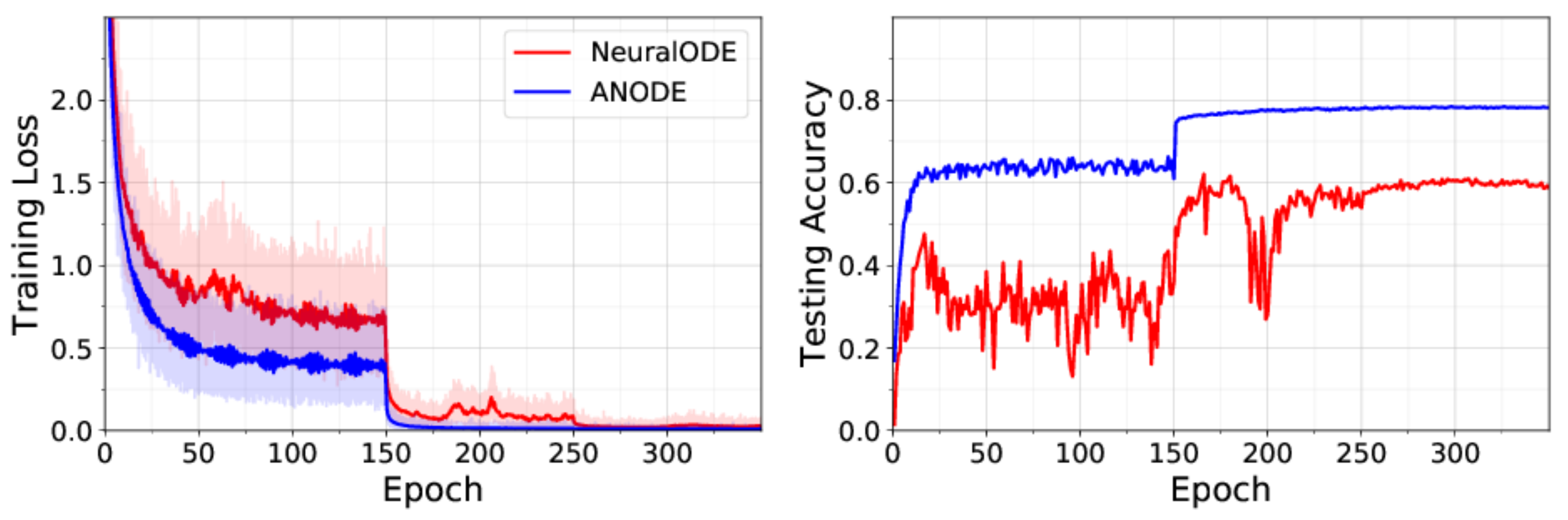}
\caption{Training loss (left) and Testing accuracy (right) for on Cifar-100.
We consider a ResNet-18 network where non-transition blocks are replaced
with an ODE block, solved with the Euler method.
As one can see, the gradient computed using~\cite{chen2018neural} results
in sub-optimal performance, compared to \OURS.
Furthermore, testing~\cite{chen2018neural} with RK45 leads to divergent training in the first epoch.
}
\label{fig:resnet_dto_vs_otd_cifar100}
\end{figure}

\begin{figure}[t]
\centering
\includegraphics[width=.49\textwidth]{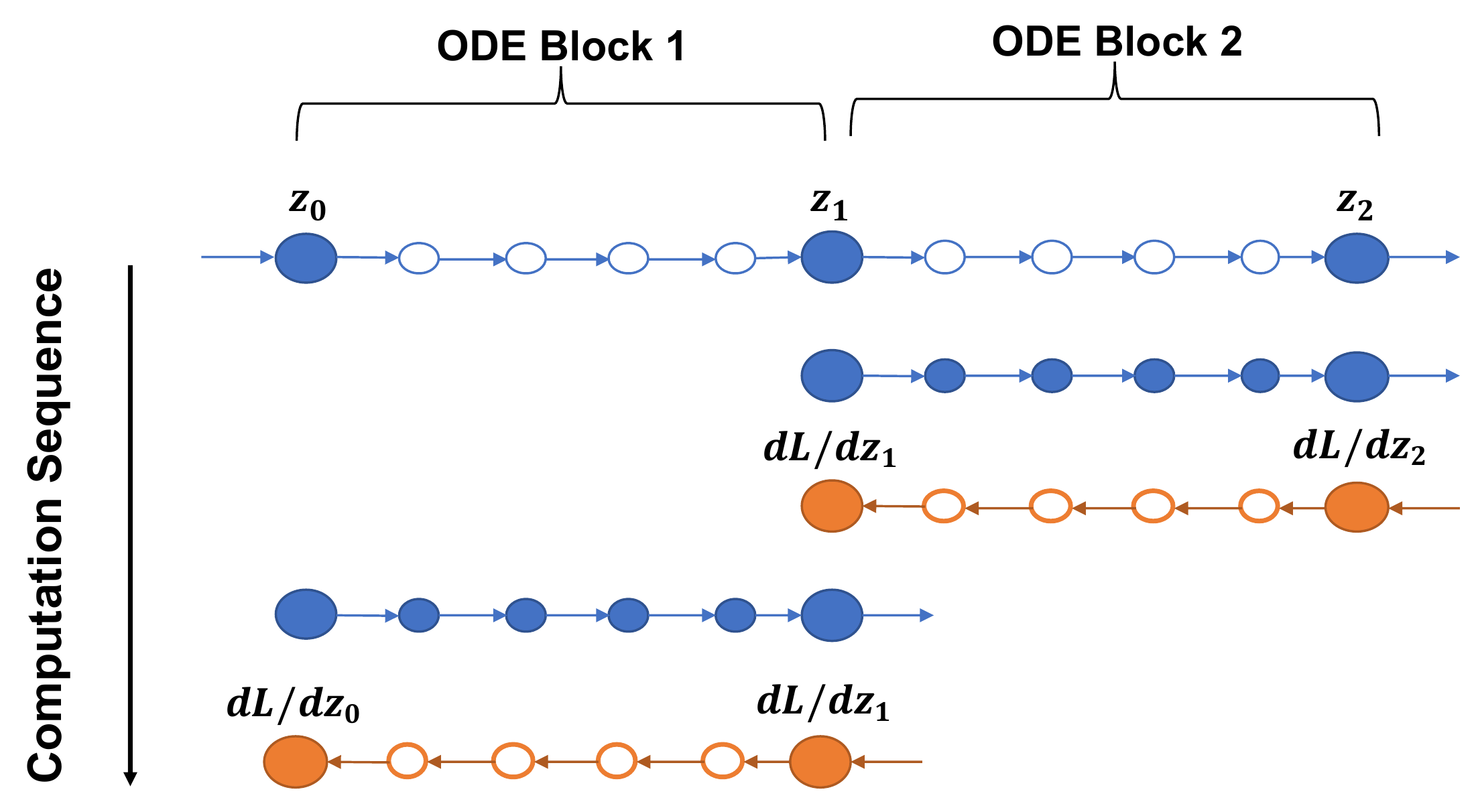}
\caption{
Illustration of checkpointing scheme for two ODE blocks
with five time steps. Each circle
denotes a complete residual block (\fref{fig:resnet_ode}).
Solid circles denote activations for an entire residual block, along
with intermediate values based on the discretization scheme,
that are stored in memory during forward pass (blue arrows).
For backwards pass, denoted by orange arrows, we first
recompute the intermediate activations of the second block,
and then solve the adjoint equations using $\frac{dL}{dz_2}$ as 
terminal condition and compute $\frac{dL}{dz_1}$.
Afterwards, the intermediate activations are freed in the memory and
the same procedure
is repeated for the first ODE block until we compute $\frac{dL}{dz_0}$.
For cases with scarce memory resources, a logarithmic checkpointing
can be used~\cite{griewank1992achieving,griewank2000algorithm}.
}
\label{fig:checkpointing}
\end{figure}

\section{\OURS}
\label{sec:sqode}

As discussed above, the main challenge with neural ODEs is the prohibitive memory
footprint during training, which has limited their deployment
for training deep models. 
For a NN with $L$ ODE layers, the memory requirement is $\bigO(LN_t)$, where $N_t$ is
the total number of time steps used for solving~\eref{e:ode}. We also need to
store intermediate values between the layers of a residual block ($z_n^i$ in~\fref{fig:resnet_ode}), which adds a constant multiplicative factor.
Here, we introduce \OURS, a neural ODE with checkpointing
that squeezes the memory footprint to $\bigO(L) + \bigO(N_t)$, with the same computational complexity as the method proposed by~\cite{chen2018neural}, and
utilizes the DTO method to backpropogate gradient information.
In \OURS, the input activations of every ODE block are stored in memory, which
will be needed for backpropogating the gradient. This amounts to the $\bigO(L)$
memory cost~\fref{fig:checkpointing}.
The backpropogation is performed in multi-stages.
For each ODE block, we first perform a forward solve on the input activation that was stored, and save intermediate results of the forward pass in memory
(i.e. trajectory of~\eref{e:ode} along with $z_n^i$ in~\fref{fig:resnet_ode}).
This amounts to $\bigO(N_t)$ memory cost. This information
is then used to solve the adjoint backwards in time using the DTO method through automatic differentiation.
Once we compute the gradients for every ODE block, this memory is released and reused for the next ODE block.
Our approach gives the correct values for ReLU activations, does not suffer from
possible numerical instability incurred by solving~\eref{e:ode} backwards in time, and
gives correct numerical gradient in line with the discretization scheme used to
solve~\eref{e:ode}.

For cases where storing $\bigO(N_t)$ intermediate activations
is prohibitive, we can incorporate the
classical checkpointing algorithm algorithms~\cite{griewank1992achieving,griewank2000algorithm}.
For the extreme case where we can only checkpoint one time step, we have
to recompute $\bigO(N_t^2)$ forward time stepping for the ODE block.
For the general case where we have $1<m<N_t$ memory available,
a naive approach would be to checkpoint the trajectory using equi-spaced
discretization. Afterwards, when the trajectory is needed for a time point
that was not checkpointed, we can perform a forward solve using the nearest
saved value.
However, this naive approach is not optimal in terms of additional
re-computations that needs to be done.
In the seminal work of~\cite{griewank1992achieving,griewank2000algorithm} an optimal strategy was proposed which carefully chooses checkpoints, such that minimum additional re-computations are needed. This approach can be directly incorporated to neural ODEs for cases with scarce memory resources. 

Finally we show results using \OURS, shown in~\fref{fig:dto_vs_otd} for a SqueezeNext network on Cifar-10 dataset. Here, every (non-transition) block of SqueezeNext is replaced with an ODE block, and~\eref{e:ode} is solved with
Euler discretization, along with an additional experiment where we use RK2 (Trapezoidal rule). As one can see, \OURS results in a stable training
algorithm that converges to higher accuracy as compared to the neural ODE method
proposed by~\cite{chen2018neural}.\footnote{We also tested~\cite{chen2018neural} with RK45 method
but that lead to divergent results.}
Furthermore, Figure~\ref{fig:resnet_dto_vs_otd} shows results using a variant of ResNet-18, where the non-transition
blocks are replaced with ODE blocks, on Cifar-10 dataset. 
Again we see a similar trend, where training neural ODEs results in sub-optimal
performance due to the corrupted gradient backpropogation.
We also present results on Cifar-100 dataset in~\fref{fig:resnet_dto_vs_otd_cifar100} with the same trend.
The reason for unconditional stability of \OURS's gradient computation is that we compute the correct gradient (DTO) to machine precision.  However, the neural ODE~\cite{chen2018comparison} does not provide any such guarantee, as it changes the calculation of the gradient algorithm which may lead to incorrect descent signal.

\section{Conclusions}
Neural ODEs have the potential to transform NNs architectures,
and impact a wide range of applications. 
There is also the promise that neural ODEs could result in models that can learn more complex tasks. Even though the link between NNs and ODEs have been known for some time, their prohibitive memory footprint has limited their deployment.  Here, we performed a detailed analysis of adjoint based methods, and the subtle issues that may arise with using such methods. In particular, we discussed the recent method proposed by~\cite{chen2018neural}, and showed that (i) it may lead to numerical instability for general convolution/activation operators, and (ii) the optimize-then-discretize approach proposed can lead to divergence due to inconsistent gradients. To address these issues, we proposed \OURS, a DTO framework for NNs which circumvents these problems through checkpointing and allows efficient computation of gradients without imposing restrictions on the norm of the weight matrix (which is required for numerical stability as we saw for~\eqref{e:matrelu}). \OURS reduces the memory footprint from $\bigO(LN_t)$ to $\bigO(L) + \bigO(N_t)$, and has the same computational cost as the neural ODE proposed by~\cite{chen2018neural}. It is also possible
to further reduce the memory footprint at the cost of additional computational overhead using classical checkpointing  schemes~\cite{griewank1992achieving,griewank2000algorithm}. We discussed results on Cifar-10/100 dataset using variants of Residual and SqueezeNext networks.

{\bf Limitations:}  A current limitation is the stability of solving~\eref{e:ode} forward in time.
\OURS does not guarantee such stability as this needs to be directly enforced either directly in the NN architecture (e.g. by using Hamiltonian systems) or implicitly through regularization. Furthermore, we do not consider NN design itself to propose a new macro architecture. However, \OURS provides the means to efficiently perform Neural Architecture Search for finding an ODE based architecture for a given task.
Another important observation, is that we did not observe generalization benefit when using more sophisticated discretization algorithms such as RK2/RK4 as compared to Euler which is
the baseline method used in ResNet. We also did not observe benefit in using more time steps as opposed to one time step. We believe this is due to staleness of the NN parameters in time.
Our conjecture is that the NN parameters need to dynamically change in time along with the activations. We leave this as part of future work.

\section*{Acknowledgments}
We would like to thank Tianjun Zhang for
proofreading the paper and providing valuable feedback.
This would was supported by a gracious fund from Intel corporation, and in particular
Intel VLAB team. We are also grateful for a gracioud fund from Google Cloud, as well as  
NVIDIA Corporation with the donation of the Titan Xp GPUs that was partially used for this research.
{ 
\printbibliography
}
\onecolumn
\appendix
\section{Appendix}
Here, we first present the derivation of the continuous Karush-Kuhn-Tucker (KKT) optimality conditions, and
then discuss the corresponding Discretize-Then-Optimize formulation.

\subsection{Extra Results}

\begin{figure}[!htbp]
\begin{center}
  \includegraphics[width=0.8\textwidth,trim=0 0 0 5in, clip]{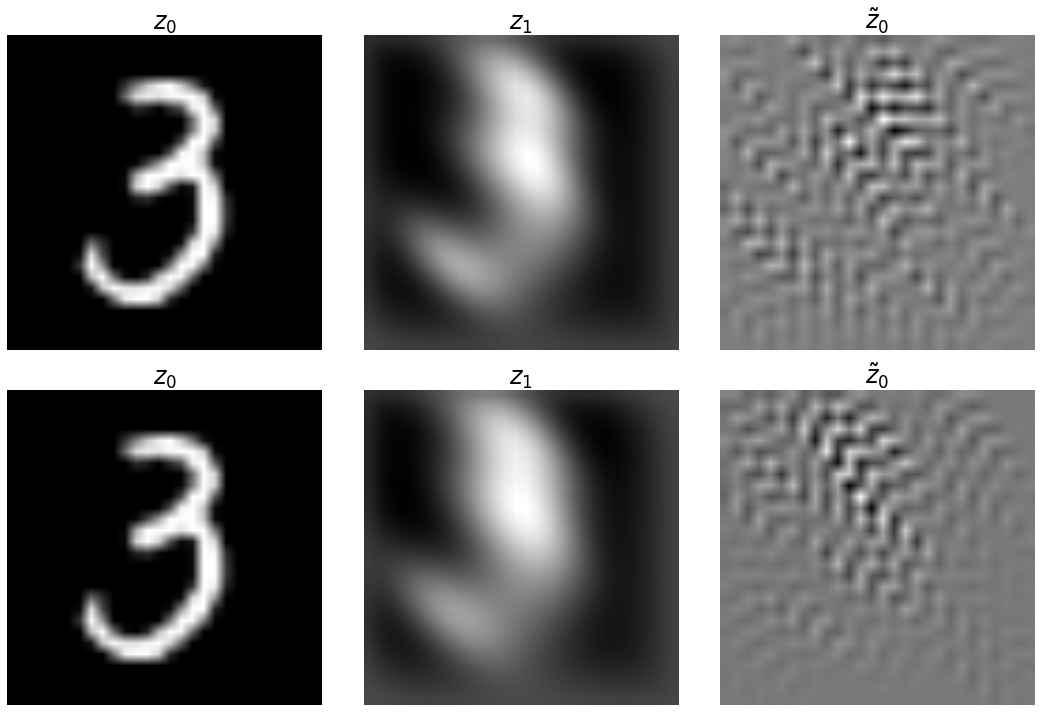}\\
  \includegraphics[width=0.8\textwidth,trim=0 0 0 5in, clip]{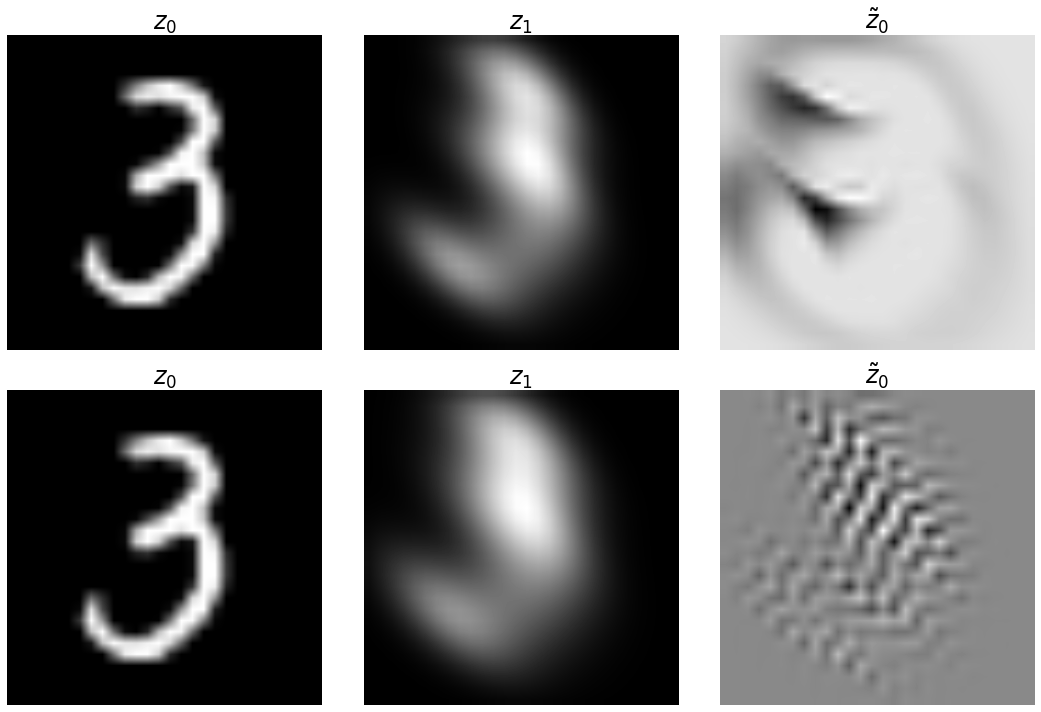}\\
  \includegraphics[width=0.8\textwidth,trim=0 0 0 5in, clip]{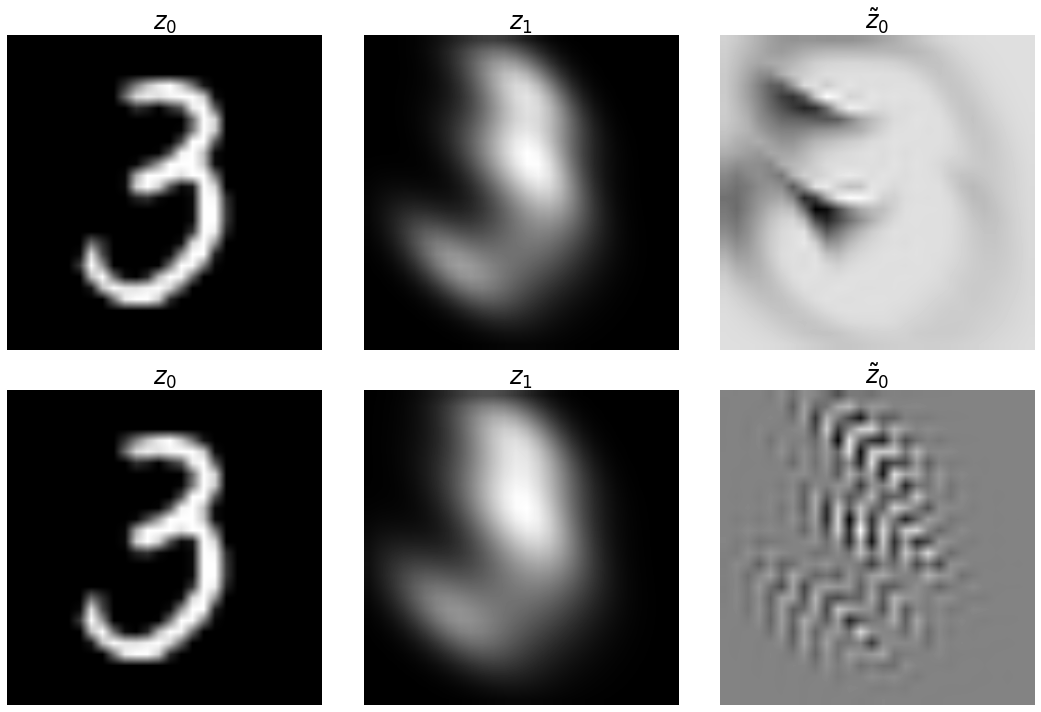}\\
  \includegraphics[width=0.8\textwidth,trim=0 0 0 5in, clip]{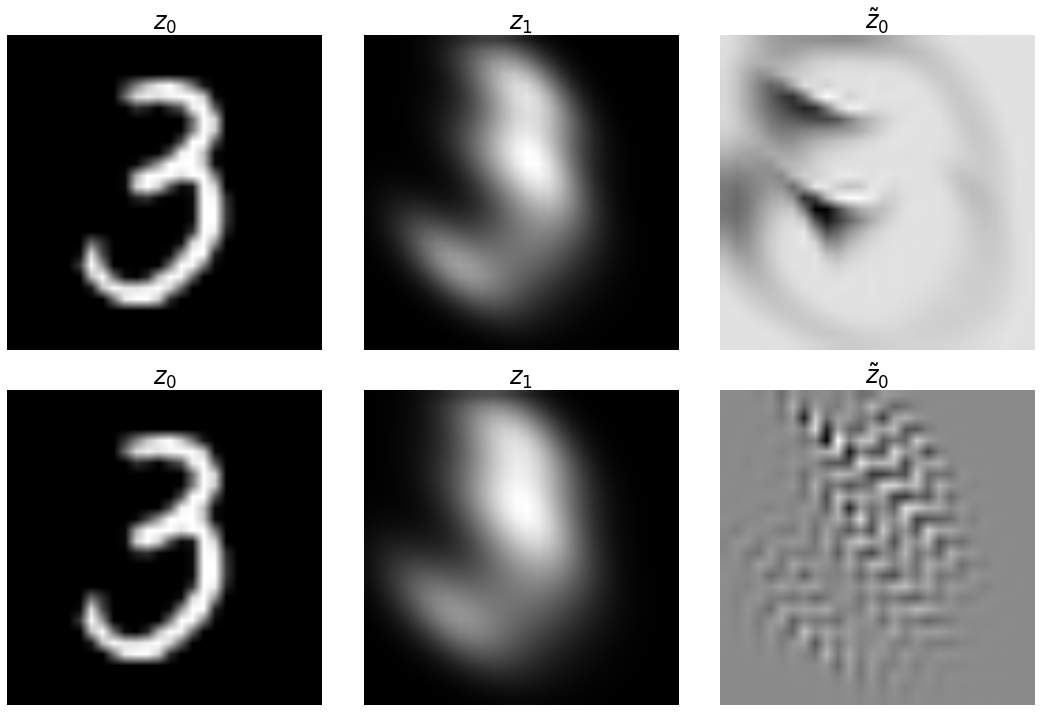}
 \end{center}
\caption{
Demonstration of numerical instability associated with reversing
NNs with adaptive RK45 solver.
We consider a single residue block with a single $3\times3$ convolution layers.
Each row corresponds to the following setting for activation after this convolution layer:
(1) no activation, (2) with ReLu activation, (3) with Leaky ReLu, and (4) with Softplus activation.
The first column shows input image that is fed to the residual block,
and the corresponding output activation is shown in the second column.
The last column shows the result when solving the forward problem backwards as proposed by~\cite{chen2018neural}.
One can clearly see that the third column is completely
different than the original image shown in the first column. This is due to the fact
that one cannot solve a general ODE backwards in time, due to numerical instabilities.
}
\label{fig:extra_results_adaptive}
\end{figure}
\subsection{Optimize Then Discretize Approach}
\label{s:OTD}

In this section, we present a detailed derivation of the optimality conditions corresponding to~\eref{e:kkt_alpha}.
We need to find the so-called KKT conditions, which can be found by finding stationary points of the corresponding Lagrangian, defined as:

\begin{align}
    \gL(z, \alpha, \theta) = \gJ(z_1, \theta) + \int_0^1\alpha(t) \cdot \left(\frac{dz}{dt} - f(z,\theta)\right)dt.
\end{align}

The optimality conditions correspond to finding the saddle point of the Lagrangian which can be found by taking variations
w.r.t. adjoint, state, and NN parameters. Taking variations w.r.t. adjoint is trivial and results in~\eref{e:kkt_z}.
By taking variations w.r.t. state variable, z, we will have:

\begin{align}
  \begin{split}
    \frac{\partial \gL}{\partial z}\hat{z} = \frac{\partial\gJ}{\partial z_1} \hat{z}_1 + \int_0^1\alpha(t) \cdot \left(\frac{d\hat{z}}{dt} - f(\hat{z},\theta)\right)dt.
  \end{split}
\end{align}

Performing integration by parts for $\frac{d\hat{z}}{dt}$ we will obtain:

\begin{align}
  \begin{split}
    \frac{\partial \gL}{\partial z}\hat{z} = \frac{\partial\gJ}{\partial z_1} \hat{z}_1 + \int_0^1 \cdot \left(-\frac{d\alpha}{dt} - \frac{\partial f(z,\theta)}{\partial z}^T\alpha\right)\hat{z}dt + \alpha_1\hat{z}_1=0 \quad \text{for all}\ \hat{z} \in (0,1].
  \end{split}
\end{align}

Note that there is no variations for $z_0$ as the input is fixed. Now the above equality (aka weak form of optimality) has to hold for all variations of $\hat{z}$ in time.
By taking variations for only $\hat{z}_1$ and setting $\hat{z} = 0$ for $z(t)\in (0,1)$ we will have:

\begin{equation}
\frac{\partial\gJ}{\partial z_1} + \alpha_1 = 0.
\end{equation}

Note that this is equivalent to~\eref{e:kkt_alpha_1}. Furthermore, taking variations by setting $\hat{z}_1 = 0$, we will have:

\begin{equation}
  -\frac{d\alpha}{dt} - \frac{\partial f(z,\theta)}{\partial z}^T\alpha = 0,
\end{equation}
which is the same as~\eref{e:kkt_alpha}. The last optimality condition can be found by imposing stationarity w.r.t. $\theta$:

\begin{align}
  \begin{split}
    \frac{\partial \gL}{\partial \theta}\hat{\theta} = \frac{\partial R}{\partial \theta} \hat{\theta} + \int_0^1 \cdot \left(- \frac{\partial f(z,\theta)}{\partial \theta}^T\alpha\right)\hat{\theta}dt =0 \quad \text{for all}\ \hat{\theta}.
  \end{split}
\end{align}

Imposing the above for all $\theta$ leads to:

\begin{equation}
  g_\theta = \frac{\partial R}{\partial \theta} \hat{\theta} - \int_0^1  \frac{\partial f(z,\theta)}{\partial \theta}^T\alpha,
\end{equation}
which is the same as~\eref{e:kkt_theta}. These equation are the so called  Optimize-Then-Discretize (OTD) form of the backpropogation.
However, using these equations could lead to incorrect gradient, as the choice of discretization creates inconsistencies between continuous form of
equations and their corresponding discrete form.
This is a known and widely studied issue in scientific computing~\cite{gholami_thesis}. The solution to this is to derive
the so called Discretize-Then-Optimize (DTO) optimality conditions as opposed to OTD which is discussed next.
It must be mentioned that by default, the auto differentiation engines automatically perform DTO. However, deriving the exact equations
could provide important insights into the backpropogation for Neural ODEs.

\subsection{Discretize Then Optimize Approach}
\label{s:DTO}

Here we show how the correct discrete gradient can be computed via the DTO approach
for an Euler time stepping scheme to solve~\eref{e:ode}. Let us assume
that the forward pass will take $N_t=n+1$ time steps:

\begin{align}
&z_0 - z_0 = 0 \nonumber\\
&z_1 - z_0 - \dt f(z_0,\theta, t_0) =0 \nonumber \\
&\vdots \nonumber \\
&z_n - z_{n-1} - \dt f(z_{n-1},\theta, t_{n-1}) =0 \nonumber \\
&z_{n+1} - z_{n} - \dt f(z_{n},\theta, t_{n}) =0 \nonumber \\
\label{e:forward_dto}
\end{align}
where $z_i$s are the intermediate solutions to~\eref{e:ode} at time points $t=i\dt$,
and $n$ denotes the $n^{\text{th}}$ time step.
Here the dashed line indicate each time step in the forward solve.
The corresponding Lagrangian is\footnote{We note a subtle point, that we are misusing the integral notation instead of a Riemann sum for integrating over spatial domain. The time domain is correctly
addressed by using the Riemann sum.}:

\begin{align*}
  &\mathcal{L} (z,\theta, \alpha) = \mathcal{J}(z_{n+1}) \\
&+\int_{\Omega} \alpha_0(z_0 - z_0)  d\Omega \\
&+\int_{\Omega} \alpha_1(z_1 - z_0 - \dt f(z_0,\theta, t_0))d\Omega \\
&\vdots\\
&+\int_{\Omega} \alpha_n(z_n - z_{n-1} - \dt f(z_{n-1},\theta, t_{n-1}))d\Omega \\
&+\int_{\Omega} \alpha_{n+1}(z_{n+1} - z_{n} - \dt f(z_{n},\theta, t_{n}))d\Omega,
\end{align*}
where $d\Omega$ is the spatial domain (for instance the resolution and channels of the activation map in a neural network), $\alpha_i$ is the adjoint variable
at time points $t=i\dt$, and $\gJ$ is the loss which depends directly on the output activation
map, $z_{n+1}$.

Now the adjoint equations (first order optimality equation) can be obtained by taking variations w.r.t. adjoint, state and inversion parameters:

\[\arraycolsep=1.4pt\def\arraystretch{2.2}
\left\{
  \begin{array}{@{}lll@{}}
\displaystyle{\frac{\partial \gL}{\partial z_i} }&=0 \quad\Rightarrow\quad \ \textrm{adjoint equations \  }\\
\displaystyle{\frac{\partial \gL}{\partial \theta}} &=0 \quad \Rightarrow\quad \ \textrm{inversion equation}\\
\displaystyle{\frac{\partial \gL}{\partial \alpha_i}} &=0  \quad\Rightarrow\quad \ \textrm{state equations \ \ \ }
  \end{array}\right.
\]

Performing this variations leads to the following ODEs for the adjoint variable:

\begin{align}
&\alpha_{n+1} - \frac{\partial \gL}{\partial z_{n+1}} = 0 \\
&\alpha_n -\alpha_{n+1} (I + \dt \frac{\partial f(z_n,\theta, t_n)}{\partial z_n}) = 0 \\
&\vdots\\
&\alpha_1 -\alpha_{2} (I + \dt \frac{\partial f(z_1,\theta, t_1)}{\partial z_1}) = 0 \\
&\alpha_0 -\alpha_{1} (I + \dt \frac{\partial f(z_0,\theta, t_0)}{\partial z_0}) = 0 \\
\label{e:adjoint_dto}
\end{align}

To get the DTO gradient, we first need to solve~\eref{e:forward_dto} for every ODE block and store intermediate
activation maps of $z_i$ for $i=0,\cdots,n+1$. Afterwards we solve~\eref{e:adjoint_dto} starting
from the last time step value of $n+1$ and marching backwards in time to compute $\alpha_0$, which is basically the gradient w.r.t. the input activation map of $z_0$.


\end{document}